%% file: main.tex
\documentclass[10pt,twocolumn,letterpaper]{article}

\usepackage[pagenumbers]{iccv} % To force page numbers, e.g. for an arXiv version
\usepackage{multirow}
\usepackage{graphicx}


\definecolor{iccvblue}{rgb}{0.21,0.49,0.74}
\usepackage[pagebackref,breaklinks,colorlinks,allcolors=iccvblue]{hyperref}

\def\paperID{9800} % *** Enter the Paper ID here
\def\confName{ICCV}
\def\confYear{2025}

\title{DeepAudio-V1:Towards Multi-Modal Multi-Stage End-to-End Video to Speech and Audio Generation}

\author{
    \textit{Haomin Zhang$^1$$^*$, Chang Liu$^{1,2}$$^*$, Junjie Zheng$^1$, Zihao Chen$^1$, Chaofan Ding$^1$, Xinhan Di$^1$} \\
    $^1$AI Lab, Giant Network.\\
    $^2$University of Trento.\\
    \small\texttt{\{zhanghaomin, liuchang, zhengjunjie, chenzihao, dingchaofan,dixinhan\}@ztgame.com}
}

\begin{document}
\maketitle
\input{sec/0_abstract}    
\input{sec/1_intro}

\input{sec/2_related}

\input{sec/3_methods}
\input{sec/4_exp}

\input{sec/5_conclusion}
{
    \small
    \bibliographystyle{ieeenat_fullname}
    \bibliography{main}
}

\end{document}

%% file: sec/0_abstract.tex
\begin{abstract}
Currently, high-quality, synchronized audio is synthesized using various multi-modal joint learning frameworks, leveraging video and optional text inputs. In the video-to-audio benchmarks, video-to-audio quality, semantic alignment, and audio-visual synchronization are effectively achieved. However, in real-world scenarios, speech and audio often coexist in videos simultaneously, and the end-to-end generation of synchronous speech and audio given video and text conditions are not well studied. Therefore, we propose an end-to-end multi-modal generation framework that simultaneously produces speech and audio based on video and text conditions. Furthermore, the advantages of video-to-audio (V2A) models for generating speech from videos remain unclear. The proposed framework, DeepAudio, consists of a video-to-audio (V2A) module, a text-to-speech (TTS) module, and a dynamic mixture of modality fusion (MoF) module. In the evaluation, the proposed end-to-end framework achieves state-of-the-art performance on the video-audio benchmark, video-speech benchmark, and text-speech benchmark. In detail, our framework achieves comparable results in the comparison with state-of-the-art models for the video-audio and text-speech benchmarks, and surpassing state-of-the-art models in the video-speech benchmark, with WER 16.57\% to 3.15\% (+80.99\%), SPK-SIM 78.30\% to 89.38\% (+14.15\%), EMO-SIM 66.24\% to 75.56\% (+14.07\%), MCD 8.59 to 7.98 (+7.10\%), MCD\_SL 11.05 to 9.40 (+14.93\%) across a variety of dubbing settings.
\end{abstract}

%% file: sec/1_intro.tex
\section{Introduction}
\label{sec:intro}
In recent years, the field of multi-modal content generation has witnessed significant advancements, particularly in video-to-audio (V2A) \cite{wang2024frieren, pascual2024masked, tian2024vidmuse} and video-to-speech (V2S) \cite{choi2023diffv2s, kefalas2023audio, mira2022svts} synthesis. These technological tools are designed to fill the gap between rich visual content and corresponding auditory experience in applications such as automated video dubbing, personalized voice assistants, and generated narration. In this work, we explore the potential coupling of these widely explored, yet independently catered tasks into a unified framework, generating both ambient audio and expressive speech from the visual input. 

\begin{figure}[t]
    \centering
    \includegraphics[width=1\linewidth]{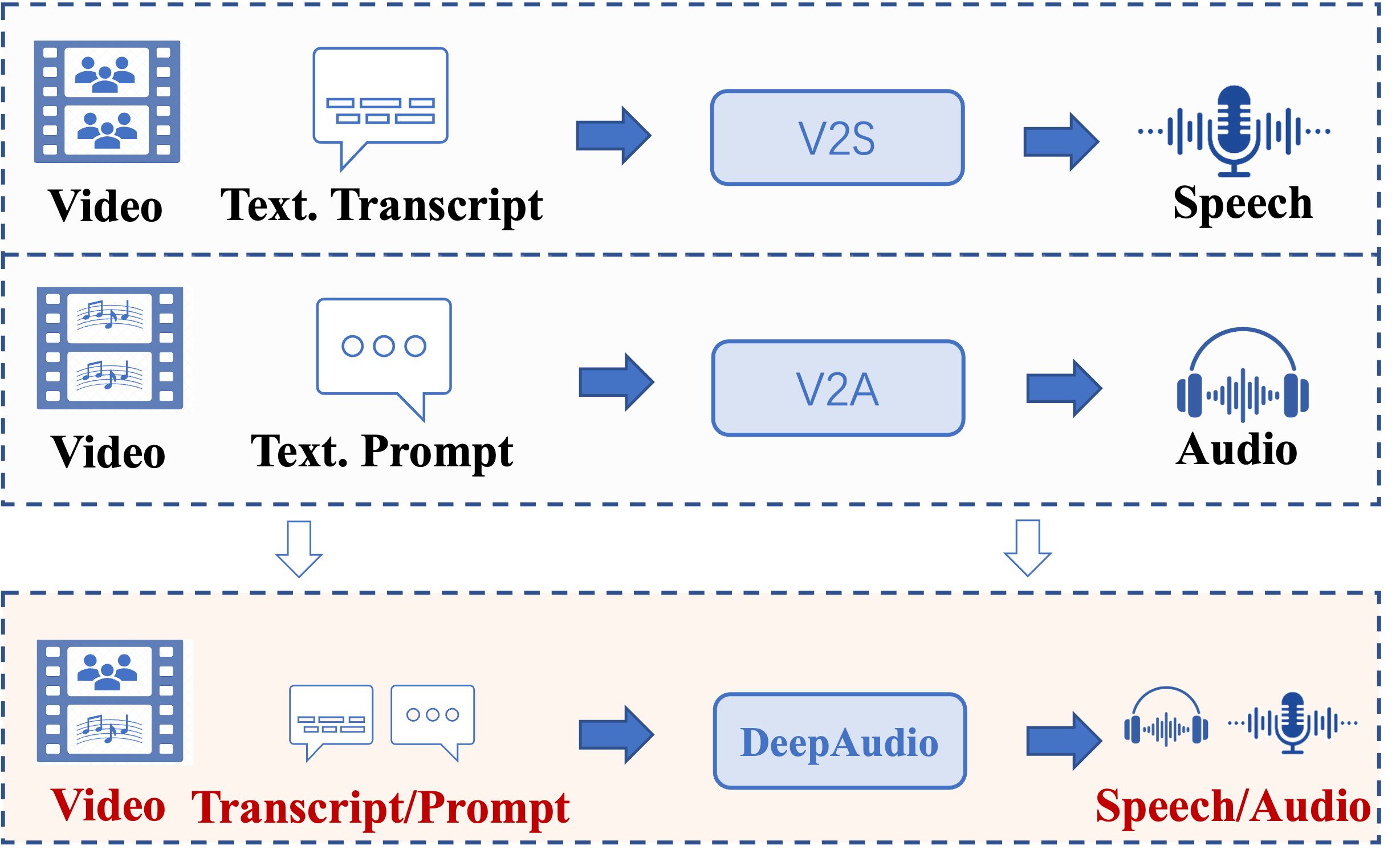}
    \caption{Overview of our DeepAudio framework. Traditional methods separately address V2A and V2S tasks. Our proposed DeepAudio framework unifies these tasks by supporting both V2A and V2S generation through one unified model, allowing users to generate either synchronized ambient sounds or expressive speech based on different textual inputs.}
    \label{fig:fig1}
\end{figure}

\subsection{Video-to-Audio Synthesis}

Video-to-audio (V2A) generation focuses on synthesizing meaningful audio that aligns with visual content. This research area has wide-ranging applications, including improving silent videos, and supporting multimedia content creation. A fundamental challenge in this domain is that visual information does not inherently carry audio signals. Instead, it provides indirect cues, such as motion, object interactions, and environmental characteristics, which should be effectively interpreted to generate realistic and contextually appropriate sounds.

Recent studies \cite{rai2024egosonics, lipman2022flow} have advanced video-driven audio synthesis by leveraging deep learning techniques to establish direct relationships between visual and auditory modalities. However, most existing approaches are tailored for specific types of audio, such as environmental noise, music, or speech, limiting their ability to generalize across different scenarios. This specialization often reduces the adaptability of such models when faced with varied and complex audiovisual environments. Consequently, a key research challenge is developing a more unified and robust framework that can dynamically generate different types of audio from visual input while maintaining both semantic coherence and temporal alignment.

\subsection{Video-to-Speech Synthesis}

Video-to-speech (V2S) generation presents a more complex challenge than general video-to-audio synthesis, as it requires generating meaningful speech that aligns with the speaker’s visual cues and surrounding context. On the basis of text-to-speech (TTS) methods, which have made significant advances in naturalness and expressiveness through deep learning techniques such as neural vocoders and transformer-based architectures~\cite{chen2024f5, du2024cosyvoice, wang2023neural, anastassiou2024seed}, V2S introduces an additional layer of complexity by relying on visual rather than textual input. Unlike traditional speech synthesis (i.e., text-to-speech), which relies on a predefined text transcript, V2S should infer speech content from visual signals. 
While recent studies have gradually made progress toward lip-synchronized speech generation, most existing methods are designed for constrained settings and find it difficult to deal with varying real-world scenarios. On the other hand, video-to-audio generation is a semantically-related problem that has seen subsequent advancements, capable of generating realistic audio from visual modalities using numerous pretrained models. However, although highly effective, such V2A models have not been readily integrated into V2S systems. At a larger scale, this would involve deploying V2S systems that leverage the knowledge of pretrained V2A models, extending beyond simple lip-reading to incorporate richer visual context, including facial expressions, gestures, and environmental interactions, ultimately generating more coherent and engaging speech.

\subsection{A Unified Multi-Modal Synthesis Framework}
To bridge the gap between V2A and V2S tasks, we propose a unified framework DeepAudio that unifies V2A and V2S generation (see Fig~\ref{fig:fig1}).Our framework consists of four-stage paradigm. (1) V2A Training: A multi-modal encoder extracts spatiotemporal features, while a synchronization module aligns generated ambient audio with visual dynamics. A flow-matching decoder refines mel-spectrograms via contrastive learning. (2) TTS Training: An emotion-controllable TTS model with prosody prediction and a flow-based decoder ensures expressive, high-fidelity speech. (3) MoF Integration: A Mixture of modality Fusion framework dynamically routes inputs, aligning cross-modal representations for efficient knowledge transfer. (4) V2S Finetuning: V2A predicts energy contours for speech synthesis, ensuring seamless audiovisual synchronization.

Our main contributions include:
\begin{itemize}
    \item \textbf{An end-to-end video to speech and audio framework} Our framework combining V2A, TTS and MoF for joint speech and audio generation, enabling end-to-end synthesis from video and text inputs.
    \item  \textbf{Knowledge-aware alignment} We pioneer the use of V2A-predicted energy contours to guide V2S synthesis.
    \item  \textbf{State-of-the-art performance} Extensive experiments on V2C-Animation \cite{chen2022v2c}, VGGSound \cite{chen2020vggsound} and LibriSpeech-PC \cite{meister2023librispeech} demonstrate SOTA results across audio-visual synchronization, speech naturalness, and emotional expressiveness metrics for V2A, V2S, and TTS datasets.
\end{itemize}

%% file: sec/2_related.tex
\section{Related Work}
\label{sec:intro}

\subsection{Text-to-Speech}
% Diffusion probabilistic models (DPMs) have demonstrated remarkable success in speech synthesis, particularly in waveform generation \cite{chen2020wavegrad} and end-to-end TTS \cite{chen2021wavegrad}. Diff-TTS \cite{jeong2021diff} first introduced DPMs for acoustic modeling, followed by Grad-TTS \cite{popov2021grad}, which employed a stochastic differential equation  formulation. While most subsequent models, such as Fast Grad-TTS \cite{vovk2022fast}, adopt a non-autoregressive paradigm, TorToiSe \cite{betker2023better} explored an autoregressive approach with quantized latent representations.

Due to its advantages of high sampling speed and generation quality, flow matching has attracted significant attention in audio generation \cite{guo2024voiceflow, kim2023p, mehta2024matcha}. Recently, Matcha-TTS \cite{mehta2024matcha} introduced optimal-transport conditional flow matching (OT-CFM) for training, which yields an ODE-based decoder to improve the fidelity of the mel spectrograms. DiTTo-TTS \cite{leeditto} improved alignment by leveraging a diffusion transformer (DiT) conditioned on text embeddings from a pretrained language model, refining neural audio codec fine-tuning. However, capturing dynamic emotional transitions remains a challenge.

More recent methods explore simplified and effective architectures. E2TTS \cite{eskimez2024e2}, based on Voicebox \cite{le2023voicebox}, removes phoneme and duration predictors, directly using padded character inputs for mel spectrograms. F5-TTS \cite{chen2024f5} further improves text-speech alignment by integrating ConvNeXt V2 \cite{woo2023convnext} into a diffusion transformer framework.

Although these models have made significant progress, most existing methods focus exclusively on text-driven speech synthesis, lacking a unified framework for video-driven speech generation. To bridge this gap, our approach integrates pretrained V2A models, enhancing V2S synthesis to enable more expressive and contextually coherent speech beyond simple lip-reading.

\subsection{Video-to-Speech}
Recent developments in V2S synthesis have focused on enhancing audiovisual alignment and speaker adaptation. Early methods such as V2C-Net \cite{chen2022v2c} and VDTTS \cite{hassid2022more} leveraged pretrained speaker embeddings \cite{jia2018transfer} to maintain identity consistency but struggled with precise lip synchronization. More recent works, such as StyleDubber \cite{cong2024styledubber} and Speaker2Dub \cite{zhang2024speaker}, introduced multiscale style adaptation and pretraining techniques to improve expressiveness, though their frame-level prosody modeling often led to unnatural articulation.

Hierarchical approaches, including HPMDubbing \cite{cong2023learning} and MCDubber \cite{zhao2024mcdubber}, improved contextual prosody through facial feature modeling and aggregation of contexts between sentences, but neglected phoneme-level pronunciation details critical for speech intelligibility. 
% Recent multimodal architectures such as Synchformer \cite{iashin2024synchformer} have advanced temporal alignment through self-supervised desynchronization detection, enabling frame-level audiovisual synchronization. 
Parallel efforts in speech-prompted adaptation \cite{wang2023neural, kharitonov2023speak} and end-to-end codec models \cite{defossez2022high, zeghidour2021soundstream} further bridged the gap between reference audio characteristics and synthesized speech, though at the cost of increased computational complexity. 

While these efforts have improved video-driven speech synthesis, most V2S models are designed solely for speech generation and do not support general audio synthesis. In contrast, our approach unifies V2S and V2A synthesis, leveraging pretrained V2A models to generate both speech and non-speech audio in a single framework, thereby enabling richer, context-aware sound generation from visual input.

\subsection{Video-to-Audio}
Recent progress in V2A synthesis has centered on three key aspects: semantic alignment, temporal synchronization, and multi-modal conditioning. To achieve semantic alignment, existing methods train on large-scale audiovisual datasets such as VGGSound \cite{chen2020vggsound} to associate visual cues with corresponding sounds \cite{wang2024v2a, huang2023make, liu2025tell}. Some models further utilize large-scale audio-text datasets like WavCaps \cite{mei2024wavcaps} to enhance cross-modal understanding through pretraining \cite{liu2024audioldm}.

Temporal synchronization has evolved from early handcrafted audio proxies, such as energy contours \cite{jeong2024read} and waveform RMS values \cite{lee2024video}, to more data-driven techniques that incorporate learned audiovisual alignment mechanisms \cite{ren2024sta}.

For multi-modal conditioning, recent approaches extend text-to-audio models by integrating visual control modules \cite{zhang2024foleycrafter, jeong2024read} or leveraging pretrained alignment scores for test-time optimization \cite{xing2024seeing}. Hybrid architectures further incorporate CLIP \cite{radford2021learning} embeddings with audio diffusion models via cross-attention layers \cite{liu2023audioldm}, enabling zero-shot adaptation to unseen object categories \cite{du2023conditional}. While some methods explore joint training across multiple modalities \cite{liu2025tell}, they often underperform compared to dedicated V2A models, underscoring the need for more effective cross-modal fusion strategies. Recent techniques address this issue through modality dropout \cite{ho2022classifier} and parameter-efficient fine-tuning of frozen foundation models \cite{tang2024codi}.

Despite significant improvements in V2A synthesis, most existing models remain specialized for general sound generation and do not explicitly support speech synthesis. Our approach bridges this gap by unifying V2A and V2S synthesis, leveraging pretrained V2A models to generate both natural environmental sounds and intelligible speech from video input, offering a more comprehensive and context-aware audiovisual synthesis framework.

%% file: sec/3_methods.tex
\section{Methods}
In order to produce high-quality audio from video, speech from video and speech from transcript in an end-to-end framework, we propose an end-to-end framework $M_{V1}$ which consists of three modules: (1) a V2A module $M_{v2a}$, (2) a TTS $M_{tts}$ module and (3) a MoF module $M_{mof}$. Particularly, the $M_{mof}$ is responsible to decide to activate the corresponding modules and parameters in the end-to-end framework to produce audio $y_{audio}^{video}$ from video, speech $y_{speech}^{video}$ from video and speech $y_{speech}^{text}$ from text according to the given instructions $x_{instruction}$. In details, the $y_{audio}^{video}$ is produced when the parameters of $M_{v2a}$ and $M_{mof}$ are activated. The $y_{speech}^{video}$ is produced when the parameters of $M_{v2a}$, $M_{tts}$ and $M_{mof}$ are activated. The $y_{speech}^{text}$ is produced when the parameters of $M_{tts}$ and $M_{mof}$ are activated. The generation process of $y_{audio}^{video}$, $y_{speech}^{video}$ and $y_{speech}^{text}$ are represented as the following:

\begin{equation}
    \begin{aligned}
        &y_{audio}^{video}|y_{speech}^{video}|y_{speech}^{text} = M_{V1}(x_{instruction}, x_{video}, \\ & x_{prompts}, x_{transcript})
    \end{aligned}
\end{equation}

In the details, and the activated parameters of the corresponding modules and the corresponding generated output are represented as the following:

\begin{equation}
    \begin{aligned}
        &y_{audio}^{video} = M_{v2a}(x_{video}, x_{prompts}) \\ &
        y_{speech}^{video} = M_{tts}(M_{mof}(x_{instruction}, M_{v2a}(x_{video}, \\& x_{prompts})), x_{transcript})\\ &
        y_{speech}^{text} = M_{tts}(M_{mof}(x_{instruction}), x_{transcript})\\
    \end{aligned}
\end{equation} 

In the following, the multi-stage training process of $M_{V1}$ and its corresponding $M_{v2a}$, $M_{tts}$ and $M_{mof}$ are represented as the following.  

\begin{figure*}
    \centering
    \includegraphics[width=1\linewidth]{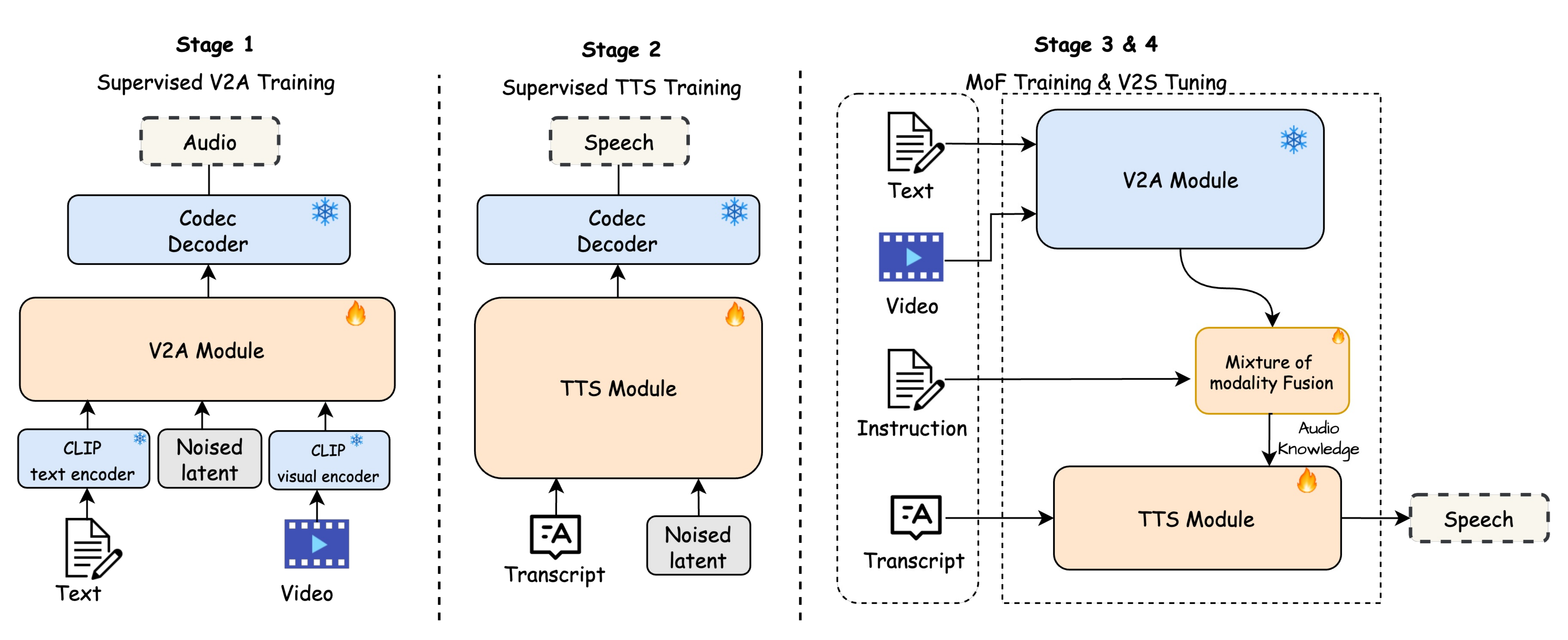}
    \caption{The overview of DeepAudio framework. The proposed DeepAudio framework unifies video-to-audio (V2A) and video-to-speech (V2S) generation in a multi-stage, end-to-end paradigm. The top section illustrates two independent generation paths: (1) The V2A module, which synthesizes ambient audio from video input using a CLIP-based multi-modal feature encoding and a noised latent representation, and (2) The TTS module, which generates speech conditioned on text and noised latent features. Both modules rely on codec decoders to reconstruct high-fidelity outputs. The bottom section presents the MoF module, an integrated multi-modal system that takes text, video, and instructions as inputs. A Gating Network adaptively fuses outputs from the V2A module and the TTS module, ensuring synchronized and context-aware audio-visual generation. }
    \label{fig:enter-label}
\end{figure*}
\subsection{Stage1: Video-to-Audio Module Learning.}

\textbf{Network Architecture}
The V2A module is designed to synthesize ambient audio that aligns accurately with visual dynamics. Our framework incorporates a multistage processing pipeline that extracts spatio-temporal features, models cross-modal synchronization, and refines mel-spectrogram representations.
The V2A module consists with two main parts, the first part $M_{v2a}^{encoder}$ works to joint learning of multi-modal feature extraction, encoding, and fusion, and the second part $M_{v2a}^{decoder}$ works to joint learning of multi-modal decoding for audio generation. Particularly, the two parts $M_{v2a}^{encoder}$ and $M_{v2a}^{decoder}$ could be a joint network architecture or two separated network architectures in an end-to-end framework.  

Formally, the V2A mechanism is defined as the following:

\begin{equation}
y_{audio}^{video} = M_{v2a}^{decoder}(M_{v2a}^{encoder}(x_{video}, x_{prompts})) 
\end{equation}

To optimize the $M_{v2a}$ network architecture, we minimize the reconstruction loss $\left[ \mathcal{L}_{\text{v2a}}(x) \right]$. 

\begin{equation}
    \min_{W_{\text{v2a}}} \mathbb{E}_{x \sim D} 
    \left[ \mathcal{L}_{\text{v2a}}(x) \right],
    \label{equ:optimization}
\end{equation}

\textbf{Flow-Matching Optimization.}
In order to achieve high-quality generated audio from the video, the flow-matching optimization is applied as this optimization method demonstrates goodness including (1)good flow matching: yielding smooth vector fields which allows for stable and efficient numerical integration (using ODE solvers), (2) reducing the number of function evaluations (NFEs) during sampling. The flow-matching optimization is applied as the following:

\begin{equation}
\mathcal{L}_{\text{FM}}(\theta) = \mathbb{E}_{t \sim \mathcal{U}(0,1),\, x_t \sim p_t(x)} \left[ \left\| f_\theta(x_t,t) - u_t(x_t) \right\|_2^2 \right]
\end{equation}

where $f_\theta(x_t,t)$ represents neural network parameterizing the learned vector field $u_t(x_t)$ represents the target (or conditional) vector field along the probability path from the source distribution to the data distribution. 

%add [overview] equation1
%add [overview] optimization equation2
%add [overview] loss equation3
%part1 [multi-modal feature extraction, encoding and fusion]
%part2 [multi-modal audio decoding and generation]
%part3 [flow-matching objective learning] 有 flow-matching 和 non flow-matching

%\textbf{Synchronization and Energy Contour Prediction} 
%The decoder alternates between multi-modal cross-attention blocks (handling visual-textual conditions) and audio-centric self-attention blocks (modeling long-range audio dependencies). This design enables joint training on both audio-visual-text triplets and audio-text pairs through modality masking, where missing modalities are replaced with learnable null embeddings during forward passes.

%\textbf{Flow-Matching Decoder with Contrastive Learning}
%For mel-spectrogram generation, we adopt a flow-matching decoder trained via contrastive learning, enhancing the realism and diversity of the synthesized audio. Unlike traditional GAN-based approaches, which may introduce artifacts, our flow-matching method ensures smooth and natural transitions between frequency components. The decoder refines spectrograms through a progressive denoising process, mitigating spectral distortions and improving clarity.

\subsection{Stage 2: Text-to-Speech Module Learning}
In the second stage of DeepAudio, the foundational speech generation model will first be trained. We use the same architecture as F5-TTS~\cite{chen2024f5}, which employs a diffusion transformer (DiT) as the backbone and is trained to output a vector field $v_\tau$ with the conditional flow-matching objective $\mathcal{L}_{CFM}$~\cite{lipman2022flow}. 
Following the supervised description stage, the training objective incorporates a conditional flow matching loss to guide the model’s learning process:

\begin{equation}
    y_{speech}^{text} = M_{tts}(x_{transcript})\\
\end{equation}
% where $M_{\text{tts}}$ represents the transformation function that generates the predicted speech representation  y_{speech}^{text} from the multimodal fused features. The function $M_{\text{mof}}$ first integrates the instructional input $x_{\text{instruction}}$ with the transcript $x_{\text{transcript}}$, producing a unified representation that enhances the speech synthesis process.
where $M_{tts}$ represents the transformation function that maps input transcript $x_{transcript}$ to the predicted speech representation $y_{speech}^{text}$.

In this step, our goal is to minimize:

\begin{equation}
\begin{aligned}
\min_{\theta} \quad & \mathbb{E}_{\tau, q(x_1), p(x|x_1)} 
\left[ \lVert u_\tau(x|x_1) - v_\tau(x; \theta) \rVert^2 \right], \\
\quad & \mathcal{L}_{\text{CFM}}(\theta) =  \lVert {y}_{speech}^{text} - speech_{gt} \rVert^2.
\end{aligned}
\end{equation}

where $p_\tau$ represents the probability path at time $\tau$, $u_\tau$ is the designated vector field for $p_\tau$, $x_1$ symbolizes the random variable corresponding to the training data, and $q$ is the distribution of the training data. The optimization ensures that the predicted trajectory $v_\tau(x; \theta)$ aligns closely with the designated flow $u_\tau(x|x_1)$, improving the overall synthesis quality. Additionally, the loss function $\mathcal{L}_{\text{CFM}}(\theta)$ ensures that the predicted speech representation ${y}_{speech}^{text}$ remains close to the ground truth speech $speech_{gt}$, enhancing both accuracy and naturalness.

\subsection{Stage 3: Dynamic Instruction Module Training.}

\begin{figure}
    \centering
    \includegraphics[width=0.8\linewidth]{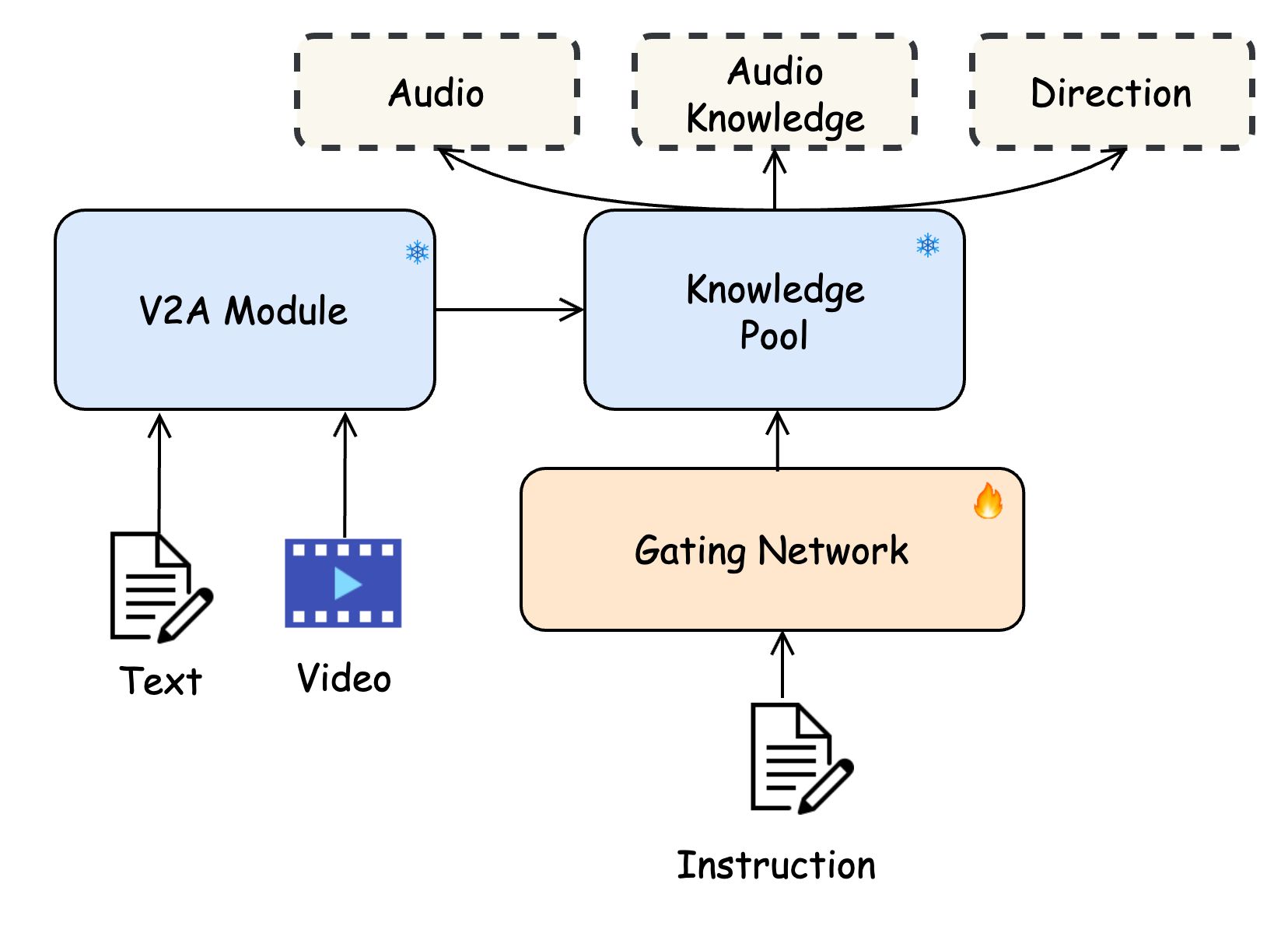}
    \caption{\textbf{Mixture of modality Fusion (MoF) framework.} The MoF framework dynamically fuses multi-modal inputs (text, video, instructions, and transcripts) through a Gating Network, which adaptively routes features to the V2A and TTS modules. This enables flexible and context-aware audio-visual generation.}
    \label{fig:fig3}
\end{figure}

% As shown in Fig.~\ref{fig:fig3}, we introduce Dynamic Instruction Training in the third stage, utilizing the Mixture of Modality Fusion (MoF) framework for unified modeling of Video-to-Audio (V2A) and Video-to-Speech (V2S) tasks. Inputs—text prompts, videos, or transcripts—are processed by the CLIP model to extract cross-modal representations, which are then passed to the Gating Network. Based on input modality and task requirements, the network adaptively assigns weights and selects feature pathways for fusion. The fused information is optimized in the Knowledge Pool to enhance representation learning for multimodal generation. Formally, the MoF mechanism is defined as:

% \begin{equation}
%     \begin{aligned}
%         F_{\text{MoF}}(x; \Theta, W_{\text{v2a}}, W_{\text{tts}}) = & G_{\text{v2a}}(x; \Theta) F_{\text{v2a}}(x; W_{\text{v2a}}) \\
%         & + G_{\text{tts}}(x; \Theta) F_{\text{tts}}(x; W_{\text{tts}}),
%     \end{aligned}
%     \label{equ:mof}
% \end{equation},
% where gating network $G(x; \Theta)_i$ is defined by:
% \begin{equation}
%     G(x; \Theta)_i = \frac{\exp(g(x; \Theta)_i)}{\sum_{j=1}^{N} \exp(g(x; \Theta)_j)},
% \end{equation}
% Finally, the optimized information from the Knowledge Pool is mapped to multimodal outputs, including Audio, Audio Knowledge, and Direction Information.

As shown in Fig.~\ref{fig:fig3}, we introduce Dynamic Instruction Training in the third stage, utilizing the MoF framework for unified modeling of V2A and V2S tasks. Tasks related instructions are passed to the Gating Network. Based on input of V2A results and instruction of task requirements, the network dynamically selects the outputs of audio, audio knowledge such as energy or pitch feature, and the direction of future tasks such as V2S. The fused information is used to enhance the next stage of multi-modal generation.

Formally, the MoF mechanism is defined as:
\begin{equation}
\resizebox{0.45\textwidth}{!}{$
\begin{aligned}
M_{mof}(x_{instruction}) =
\\ & G_{v2s}(x_{instruction}) \cdot M_{v2a}(x_{video}, x_{prompts}) 
\\ & + G_{tts}(x_{instruction}) \cdot \phi,
\end{aligned}
$}
\label{equ:mof}
\end{equation}

where $M_{mof}$ represents the final output feature representation, the Audio Knowledge in \ref{fig:fig3}, $M_{v2a}$ denotes the V2A module from Stage 1-2, $\phi$ is an empty feature, and $G_{v2s}$, $G_{tts}$ are the gating weights for V2S and TTS tasks respectively, the task Direction in \ref{fig:fig3}.

We optimize the Gating Network using a multi-class Cross-Entropy loss function $\mathcal{L}_{\text{Gating}}(x)$, which is defined as:

\begin{equation}
\resizebox{0.45\textwidth}{!}{$
    \begin{aligned}
    \mathcal{L}_{\text{Gating}} & = - \frac{1}{N} \sum_{n=1}^{N} \sum_{k=1}^{K} y_{nk} \log(p_{nk}),
    \\ p_{nk} = G(x_{instruction}) & = MLP(TextEncoder(x_{instruction}))
    \end{aligned}
$}
\end{equation}

Here, $p_{nk}$ represents the predicted probability of sample $n$ to be in class $k$, $y_{nk}$ represent whether sample $n$ belongs to class $k$, 1 or 0. We simply use MLP to do the prediction and FLAN-T5~\cite{chung2024scaling} to encode human instructions.

Finally, the optimized information from the Knowledge Pool is mapped to multi-modal outputs, including Audio, Audio Knowledge, and Direction Information. This ensures that the generated outputs maintain high fidelity and relevance to the given input context, leveraging cross-modal interactions for improved synthesis quality.
\subsection{Stage 4: Video-to-Speech Module Tuning.}
\begin{figure}
    \centering
    \includegraphics[width=0.5\linewidth]{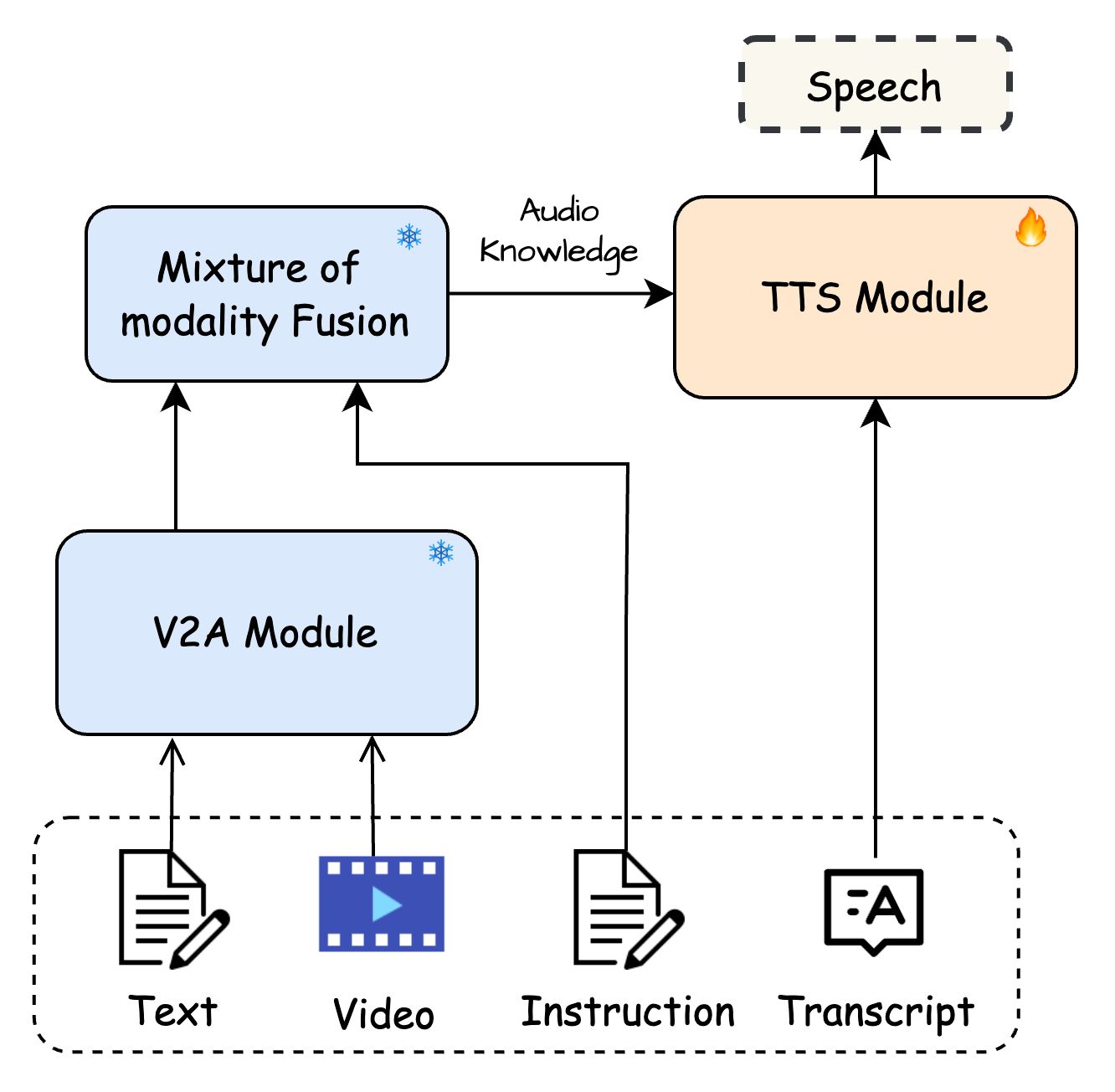}
    \caption{\textbf{Video-to-Speech Tuning.} The V2A-predicted energy contours and transcripts guide the TTS module to generate synchronized speech, ensuring improved alignment with video input through cross-modal conditioning.}
    \label{fig:fig4}
\end{figure}

%in the stage, we then use the outcome of dynamic instruction module as the condition to tuning TTS module which is trained in stage 2.

% Next, in the video-to-speech tuning stage, the energy contours come from V2A module, along with transcript are fed as inputs to the speech generation model. In this context, the provided energy contours helps guide the video-to-speech generation process, as shown in Figure \ref{fig:fig4}. The proposed speech generation model takes as input the script, silent video, and reference speech, and generates video-aligned speech context sequences, which can be described as:
% \begin{equation}
% \hat{S} = F_{generator}(V_l, energy_{contours}, T_v)
% \end{equation} 
% \noindent where $V_l$ represents the visual features derived from the input video frames, \(T_v\) represents the embedded script, $energy_{contours}$ represents the energy contour predicted by the V2A module.
% We implement a cross-attention mechanism that facilitates the integration of understanding conclusion features $C_v$ and visual features $V_l$. 
% The final loss function is constructed as follows:
% \begin{equation}
%     \resizebox{.9\linewidth}{!}{$
%     \mathcal{L}_{\text{g}} = \mathbb{E}_{\tau, q(x_1), p(x|x_1)} \lVert u_\tau(x|x_1, v_l, ey_v, t_v) - v_\tau(x; \theta) \rVert^2 
% $}
% \end{equation}
% In the training stage, the visual condition $V_l$, video-guided energy contours $ey_v$, and video script condition $T_v$ are each set to $\phi$ with a 5 \% probability.
Next, in the V2S tuning stage, the energy contours come from the V2A module, along with the transcript, and are fed as inputs to the speech generation model. In this context, the energy contours provided, which are audio knowledge we used, help guide the V2S generation process, as shown in Figure \ref{fig:fig4}. The proposed speech generation model takes as input the script, silent video, and reference speech, and generates video-aligned speech context sequences, which can be described as:

\begin{equation}
    \begin{aligned}
        &y_{speech}^{video} = M_{tts}(M_{mof}(x_{instruction}, M_{v2a} 
             (x_{video}, \\& x_{prompts})), x_{transcript})\\ &
    \end{aligned}
\end{equation}

where $x_{\text{video}}$ represents the visual features derived from the input video frames, $x_{\text{transcript}}$ represents the input transcript, and $\text{energy}_{\text{contours}}$ represents the energy contour predicted by the V2A module.

% \subsubsection{Optimization Strategy}

The optimization strategy aims to ensure that the predicted feature trajectory remains aligned with the expected V2S transformation. The optimization process is formulated as follows:

\begin{equation}
\begin{aligned}
    &\min_{\theta} \mathbb{E}_{\tau, q(x_1), p(x|x_1)} 
    \left[ \lVert u_\tau(x|M_{mof}(x_{\text{instruction}}, \right. \\
    &\phantom{=}\quad M_{\text{v2a}}(x_{\text{video}}, x_{\text{prompts}})), x_{\text{transcript}})\phantom{=}\left. - v_\tau(x; \theta) \rVert^2 \right],
\end{aligned}
\end{equation}

    where: $\\
\begin{aligned}
    u_\tau(x | M_{mof}(x_{\text{instruction}}, M_{\text{v2a}}(x_{\text{video}}, x_{\text{prompts}})), 
    x_{\text{transcript}})
\end{aligned}
$
represents the gating network of MoF, energy contours from V2A module, and the transcript. The goal is to minimize the deviation between the predicted transformation $v_\tau(x; \theta)$ and the designated feature trajectory $u_\tau$.

To further improve model robustness, during training, each of the conditioning variables $M_{mof}$, and $x_{\text{transcript}}$ and prompt audio is randomly dropped with 25\% probability. This encourages the model to generate coherent speech and have the model learn more with text and video alignments.

% \subsubsection{Loss Function}

To further refine the generated speech quality, we employ a Mean Squared Error (MSE) loss:

\begin{equation}
    \mathcal{L} = \lVert y_{speech}^{video} - S_{\text{gt}} \rVert^2,
\end{equation}

where $S_{\text{gt}}$ is the representation of the ground truth speech, and $y_{speech}^{video}$ is the generated one. This ensures that the predicted speech closely matches the reference.
Finally, the learned representation is mapped to the final speech waveform, ensuring both temporal coherence and expressive alignment with the visual input.

%% file: sec/4_exp.tex
\section{Experimental Results}
\subsection{Implementation Details}
To ensure stable training and enhance multi-modal sound generation, we adopt a multi-stage training strategy with optimized settings for each stage. The training process is conducted on 8 NVIDIA A800 GPUs.

V2A Module: Trained with a batch size of 128 for 330k updates. We use the Adam optimizer with a declaying learning rate from 1e-4 to 1e-6, and a linear warmup schedule of 1K updates. We clip the gradient norm to 0.2 to improve stability. We trained two types of flow matching V2A methods: one based on MMAudio, using CLIP~\cite{radford2021learning} as text and video encoder, and the other based on YingSound, using FLAN-T5~\cite{chung2024scaling} as the text encoder instead.

TTS Module: Trained with a batch size of 307,200 audio frames for 1.2M updates, using the AdamW optimizer with a peak learning rate of 7.5e-5. The learning rate is linearly warmed up over 20K updates before decaying linearly. We set the maximum gradient norm clip to 1.0.

MoF Module: Trained with a batch size of 32 for 5k updates. We use the Adam optimizer with a declaying learning rate from 1e-4 to 1e-6, and a linear warmup schedule of 1K updates.

V2S Tuning: Finetuned on V2S task for 15k updates. We follow the training strategies of the TTS Module, except using a learning rate of 1e-5 with a linear warmup schedule of 1k updates.

\subsection{Datasets and Metrics}

\begin{table*}[t]
\centering
\resizebox{\textwidth}{!}{%
\begin{tabular}{lcccccc|ccccc|cc}
\toprule

\multirow{2}{*}{V2S Task} & & \multicolumn{5}{c|}{V2C-Animation Dub 1.0} & \multicolumn{5}{c|}{V2C-Animation Dub 2.0} & \multicolumn{2}{c}{V2C-Animation Dub 3.0} \\
\cmidrule(lr){3-7} \cmidrule(lr){8-12} \cmidrule(lr){13-14}
& Params & WER(\%)$\downarrow$ & SPK-SIM(\%)$\uparrow$ & EMO-SIM(\%)$\uparrow$ & MCD$\downarrow$ & MCD\_SL$\downarrow$
& WER(\%)$\downarrow$ & SPK-SIM(\%)$\uparrow$ & EMO-SIM(\%)$\uparrow$ & MCD$\downarrow$ & MCD\_SL$\downarrow$
& WER(\%)$\downarrow$ & SPK-SIM(\%)$\uparrow$ \\
\midrule
GT & & 16.10 & 100 & 100 & 0 & 0 & 16.10 & 100 & 100 & 0 & 0 & 10.19 & 100 \\
HPMDubbing~\cite{cong2023learning} & 69M & 151.02 & 73.64 & 39.85 & 8.59 & 8.32 & 150.83 & 73.01 & 34.69 & \textbf{9.11} & 12.15 & 126.85 & 68.14 \\
Speaker2Dub~\cite{zhang2024speaker} & 160M & 31.23 & 82.15 & 65.92 & 10.68 & 11.21 & 31.28 & 79.53 & 59.71 & 11.16 & 11.70 & 16.57 & 76.10 \\
StyleDubber~\cite{cong2024styledubber} & 163M & 27.36 & 82.48 & 66.24 & 10.06 & 10.52 & 26.48 & 79.81 & 59.08 & 10.56 & 11.05 & 19.07 & 78.30 \\
Ours-MM-S16K & 493M & 7.16 & 89.20 & 75.27 & 8.02 & 8.11 & 10.29 & 83.83 & 65.70 & 9.33 & 9.43 & 3.17 & 89.30 \\
Ours-MM-S44K & 493M & \textbf{6.90} & \textbf{89.22} & \textbf{75.56} & \textbf{7.98} & \textbf{8.07} & 10.46 & 83.83 & 65.65 & 9.30 & \textbf{9.40} & 3.15 & 89.38 \\
Ours-MM-M44K & 957M & 7.24 & 89.12 & 75.37 & 8.01 & 8.10 & \textbf{10.24} & 83.78 & \textbf{65.83} & 9.31 & 9.41 & 3.15 & \textbf{89.39} \\
Ours-MM-L44K & 1.37B & 7.18 & 89.19 & 75.41 & \textbf{7.98} & \textbf{8.07} & 10.40 & \textbf{83.87} & 65.70 & 9.31 & 9.42 & \textbf{3.03} & 89.22 \\
Ours-YS-24K & 1.05B & 7.33 & 89.14 & 75.49 & 8.01 & 8.10 & 10.58 & 83.84 & \textbf{65.83} & 9.32 & 9.43 & 3.10 & 89.21 \\

\midrule
\multirow{2}{*}{V2A Task} & & \multicolumn{6}{c}{VGGSound-Test} \\
\cmidrule(lr){3-8}
& Params & FAD$\downarrow$ & FD$\downarrow$ & KL$\downarrow$ & IS$\uparrow$ & \multicolumn{1}{c}{CLIP$\uparrow$} & AV$\uparrow$ \\
\cmidrule(lr){1-8}
Diff-Foley~\cite{luo2023diff} & 859M & 6.05 & 23.38 & 3.18  & 10.95 & \multicolumn{1}{c}{9.40}  & 0.21 \\
FoleyCrafter w/o text~\cite{zhang2024foleycrafter} & 1.22B & 2.38 & 26.70 & 2.53 & 9.66 & \multicolumn{1}{c}{15.57} & \textbf{0.25} \\
FoleyCrafter w. text~\cite{zhang2024foleycrafter} & 1.22B & 2.59 & 20.88 & 2.28 & 13.60 & \multicolumn{1}{c}{14.80} & 0.24 \\
V2A-Mapper~\cite{wang2024v2a} & 229M & 0.82 & 13.47 & 2.67 & 10.53 & \multicolumn{1}{c}{15.33} & 0.14 \\
Frieren~\cite{wang2024frieren} & 159M & 1.36 & 12.48 & 2.75 & 12.34 & \multicolumn{1}{c}{11.57} & 0.21 \\
Ours-MM-S16K & 157M & 0.79 & 5.22 & \textbf{1.65} & 14.44 & \multicolumn{1}{c}{15.22} & 0.21 \\
Ours-MM-S44K & 157M & 1.66 & 5.55 & 1.67 & \textbf{18.02} & \multicolumn{1}{c}{15.38} & 0.21 \\
Ours-MM-M44K & 621M & 1.13 & 4.74 & 1.66 & 17.41 & \multicolumn{1}{c}{15.89} & 0.22 \\
Ours-MM-L44K & 1.03B & 0.97 & \textbf{4.72} & \textbf{1.65} & 17.40 & \multicolumn{1}{c}{16.12} & 0.22 \\
Ours-YS-24K & 718M & \textbf{0.74} & 5.69 & 1.69 & 14.63 & \multicolumn{1}{c}{\textbf{17.70}} & 0.24 \\

\cmidrule(lr){1-8}
\multirow{2}{*}{TTS Task} & & \multicolumn{2}{c}{LibriSpeech-PC test-clean} \\
\cmidrule(lr){3-4}
 & Params & SIM-O$\uparrow$ & WER(\%)$\downarrow$ \\
\cmidrule(lr){1-4}
GT & & 0.665 & 2.281 \\
F5-TTS~\cite{yushenf5} & 336M & 0.648 & 2.508 \\
Ours-MM-S16K & 336M & \textbf{0.650} & 2.267 \\
Ours-MM-S44K & 336M & 0.648 & 2.499 \\
Ours-MM-M44K & 336M & 0.648 & 2.510 \\
Ours-MM-L44K & 336M & 0.647 & \textbf{2.175} \\
Ours-YS-24K & 336M & 0.647 & 2.521 \\

\bottomrule
\end{tabular}}
\caption{Comparison of different methods on V2S, V2A and TTS tasks. Ours-MM-* represents the V2A module is a reproduced MMAudio~\cite{cheng2024taming} series model with specific parameter size and audio sampling rate, while Our-YS-24K represents the V2A module is a reproduced YingSound~\cite{chen2024yingsound} model. The Params of the Ours-* model refers to the number of actually activated parameters, which depends on the specific task.}
\label{tab:v2c_v2a_tts}
\end{table*}

\begin{table*}[t]
\centering
\resizebox{\textwidth}{!}{%
\begin{tabular}{lccccc|ccccc|cc}
\toprule

\multirow{2}{*}{V2S Task} & \multicolumn{5}{c|}{V2C-Animation Dub 1.0} & \multicolumn{5}{c|}{V2C-Animation Dub 2.0} & \multicolumn{2}{c}{V2C-Animation Dub 3.0} \\
\cmidrule(lr){2-6} \cmidrule(lr){7-11} \cmidrule(lr){12-13}
& WER$\downarrow$ & SPK-SIM$\uparrow$ & EMO-SIM$\uparrow$ & MCD$\downarrow$ & MCD\_SL$\downarrow$
& WER$\downarrow$ & SPK-SIM$\uparrow$ & EMO-SIM$\uparrow$ & MCD$\downarrow$ & MCD\_SL$\downarrow$
& WER$\downarrow$ & SPK-SIM$\uparrow$ \\
\midrule
MMAudio-L-44K~\cite{cheng2024taming} & 119.08 & 68.26 & 45.12 & 12.78 & 12.95 & 119.08 & 68.26 & 45.12 & 12.78 & 12.95 & 126.54 & 58.95 \\
F5-TTS~\cite{yushenf5} & 12.66 & 87.18 & 72.81 & 8.33 & 10.61 & 23.25 & 80.90 & 61.47 & 9.54 & 19.88 & 7.03 & 88.62 \\
Ours-MM-L44K & \textbf{7.18} & \textbf{89.19} & \textbf{75.41} & \textbf{7.98} & \textbf{8.07} & \textbf{10.40} & \textbf{83.87} & \textbf{65.70} & \textbf{9.31} & \textbf{9.42} & \textbf{3.03} & \textbf{89.22} \\

\bottomrule
\end{tabular}}
\caption{Ablation study on V2C-Animation benchmark.}
\label{tab:ablation}
\end{table*}

We train and evaluate our DeepAudio framework mainly on three datasets for different subtasks: V2C-Animation~\cite{chen2022v2c} for V2S, VGGSound~\cite{chen2020vggsound} for V2A, and EMILIA~\cite{he2024emilia} for TTS.
\begin{itemize}
    \item \textbf{V2C-Animation \cite{chen2022v2c}}: A specialized dataset for animated movie dubbing with 10,217 clips from 26 films, split into 60\% train, 10\% validation, and 30\% test.

    \item \textbf{VGGSound \cite{chen2020vggsound}}: A large-scale audiovisual dataset with 310 categories and ~500 hours of video. We also use about 1.2M text-audio pairs, including AudioCaps~\cite{kim2019audiocaps}, WavCaps~\cite{mei2024wavcaps}, TangoPromptBank~\cite{ghosal2023tango}, MusicCaps~\cite{agostinelli2023musiclm}, and AF-AudioSet~\cite{gemmeke2017audio,kong2024improving}, along with VGGSound for V2A training.

    \item \textbf{Emilia \cite{he2024emilia}}: A multilingual speech dataset with about 101k hours of speech. We use the English and Chinese subset (95k hours) for TTS pretraining. We evaluate on the LibriSpeech-PC test-clean dataset, following the evaluation protocol of F5-TTS~\cite{chen2024f5}.
\end{itemize}

To comprehensively evaluate our framework, we use metrics that measure semantic alignment, temporal synchronization, audio quality, speech naturalness, and speaker identity preservation.

\textbf{Video-to-Speech and Text-to-Speech Metrics}
SIM-O (Speaker Similarity) \cite{chen2022large} \& SPK-SIM \cite{cong2024styledubber}: Speaker embedding similarity.
WER (Word Error Rate): Pronunciation accuracy, and Whisper-large-v3 \cite{radford2023robust} for English.
EMO-SIM \cite{ye2023temporal}: Emotion similarity via pre-tained speech emotion recognition model.
MCD \& MCD-SL \cite{battenberg2020location}: Spectral and length differences.

\textbf{Video-to-Audio Metrics}
FAD (Fréchet Audio Distance) \& FD (Fréchet Distance) \cite{heusel2017gans}: perceptual similarity and audio quality.
KL-softmax \cite{iashin2021taming}: semantic consistency.
IS (Inception Score) \cite{salimans2016improved}: diversity and quality of generated audio.
CLIP Score: cross-modal alignment between visual and audio features.
AV-align (AV) \cite{yariv2024diverse}: synchronization between visual and audio content.

\subsection{Main Results}

\textbf{Results on Video-to-Speech} Table \ref{tab:v2c_v2a_tts} presents the objective evaluation results for the V2S task on the V2C-Animation dataset across three different test settings. For the Dub 1.0 setting, we use ground truth audio as reference audio. For the Dub 2.0 setting, we use non-ground truth audio from the same speaker within the dataset as reference audio, which better aligns with practical dubbing usage. For the Dub 3.0 setting, we use audio from unseen speakers (from another dataset) as reference audio.

Benefiting from the strong capabilities of both the V2A and TTS modules, our method achieves better performance across all metrics.
Ours-MM-S44K outperforms with WER 27.36\% to 6.90\% (+74.78\%), SPK-SIM 82.48\% to 89.22\% (+8.17\%), EMO-SIM 66.24\% to 75.56\% (+14.07\%), MCD 8.59 to 7.98 (+7.10\%), MCD\_SL 8.32 to 8.07 (+3.00\%) on Dub 1.0 setting, WER 26.48\% to 10.46\% (+60.50\%), SPK-SIM 79.81\% to 83.83\% (+5.04\%), EMO-SIM 59.71\% to 65.65\% (+9.95\%), MCD\_SL 11.05 to 9.40 (+14.93\%) on Dub 2.0 setting, WER 16.57\% to 3.15\% (+80.99\%), SPK-SIM 78.30\% to 89.38\% (+14.15\%) on Dub 3.0 setting.
Other model variants of our framework also achieve similar improvements.

\begin{figure*}
    \centering
    \includegraphics[width=0.9\linewidth]{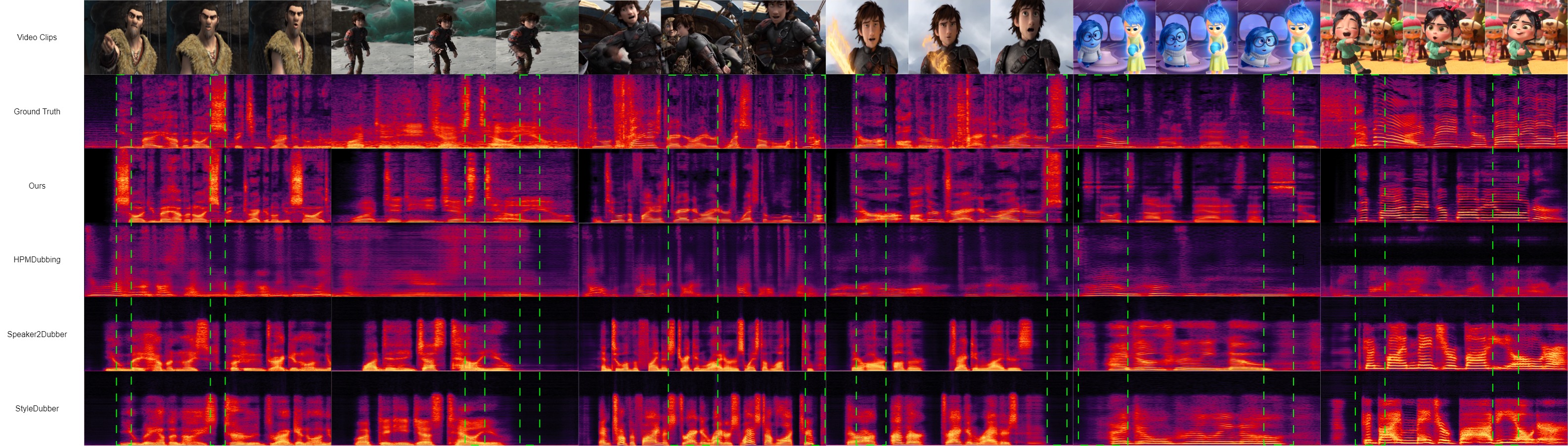}
    \caption{Mel-spectrograms of ground truth and synthesized audio samples from different methods under V2C-Animation Dub 2.0 setting.}
    \label{fig:specs}
\end{figure*}

We also present a comparison of synthesized audio generated by different approaches~\ref{fig:specs}. Spectrograms indicate that our method achieves better keypoint alignment and higher audio quality.

\textbf{Results on Video-to-Audio}
We reproduce MMAudio~\cite{cheng2024taming} and YingSound~\cite{chen2024yingsound}, and achieve similar results on VGGSound, as shown in Table ~\ref{tab:v2c_v2a_tts}.

\textbf{Results on Text-to-Speech}
Even after fine-tuning on the V2S dataset, our TTS performance metrics remain comparable to the pretrain model. Our model maintains low Word Error Rate (WER), indicating high intelligibility and pronunciation accuracy. The high Speaker Similarity (SIM-O) score shows that our approach effectively preserves speaker identity, performing on par with dedicated text-to-speech models such as F5-TTS~\cite{chen2024f5}.

\subsection{Ablation Studies}
We conduct two ablation experiments to demonstrate the effectiveness of our proposed method in Table~\ref{tab:ablation}.

By comparing \textit{MMAudio-L-44K} and \textit{Ours-MM-L44K}, it is evident that a standalone V2A model fails to generate intelligible speech, as indicated by a significant deterioration in all V2S metrics.

By comparing \textit{F5-TTS} and \textit{Ours-MM-L44K}, it is evident that due to the absence of visual information guidance, the TTS model also experiences a performance decline in the V2S task, particularly in terms of audio similarity and temporal alignment.
Our method outperforms zero-shot F5-TTS with WER 12.66\% to 7.18\% (+43.29\%), SPK-SIM 87.18\% to 89.19\% (+2.31\%), EMO-SIM 72.81\% to 75.41\% (+3.57\%), MCD 8.33 to 7.98 (+4.20\%), MCD\_SL 10.61 to 8.07 (+23.94\%) on Dub 1.0 setting, WER 23.25\% to 10.40\% (+55.27\%), SPK-SIM 80.90\% to 83.87\% (+3.67\%), EMO-SIM 61.47\% to 65.70\% (+6.88\%), MCD 9.54 to 9.31 (+2.41\%), MCD\_SL 19.88 to 9.42 (+52.62\%) on Dub 2.0 setting, WER 7.03\% to 3.03\% (+56.90\%), SPK-SIM 88.62\% to 89.22\% (+0.68\%) on Dub 3.0 setting.

%% file: sec/5_conclusion.tex
\section{Conclusion}

In this paper, we introduced DeepAudio-V1, a unified end-to-end framework for multi-modal V2A, TTS, and V2S synthesis. Unlike existing methods that separately address V2A, TTS and V2S tasks, our approach integrates these modalities into a single end-to-end framework, enabling the seamless generation of synchronized ambient audio and expressive speech from video and text inputs. To achieve these, we proposed a four-stage training pipeline that includes V2A learning, TTS learning, a dynamic MoF module, and V2S finetuning. We are in progress of developing V2 version for diverse real-world scenarios with multi-modal chain-of-thoughts guidance.
%By leveraging V2A models to guide speech synthesis, our approach improves audiovisual synchronization and speech expressiveness while maintaining high fidelity. Extensive experiments on V2C-Animation, VGGSound, and LibriSpeech-PC benchmarks demonstrate that DeepAudio achieves state-of-the-art performance across multiple evaluation metrics, including speech naturalness, speaker similarity, emotional expressiveness, and audio-visual synchronization. Ablation studies further confirm the effectiveness of integrating video-conditioned energy contours into speech synthesis, improving alignment and coherence.
%Future research directions include expanding DeepAudio to more diverse real-world scenarios, improving robustness in unseen conditions, and incorporating personalized voice adaptation for enhanced speaker consistency. We believe this work represents a significant step toward more realistic and context-aware audiovisual generation.